\pdfoutput=1

\documentclass[11pt]{article}

\usepackage[]{acl} 
\usepackage{times}       
\usepackage[T1]{fontenc} 
\usepackage[utf8]{inputenc} 
\usepackage{microtype}   
\usepackage{inconsolata}
\usepackage{graphicx}    
\usepackage{amsmath}     
\usepackage{algorithm}   
\usepackage{algpseudocode}
\usepackage{subfigure}   
\usepackage{booktabs}    
\usepackage{xcolor}      
\usepackage{amsfonts}    
\usepackage{nicefrac}    
\usepackage{hyperref}    
\usepackage{url}         
\usepackage{wrapfig}     
\usepackage{multirow}    
\usepackage{newunicodechar} 
\usepackage[capitalize,noabbrev]{cleveref} 
\usepackage{enumitem}
\definecolor{darkblue}{rgb}{0, 0, 0.5}
\definecolor{linkcolor}{rgb}{0,0.2,0.6} 

\hypersetup{
  colorlinks = true,
  linkcolor  = linkcolor,
  citecolor  = linkcolor,
  urlcolor   = linkcolor
}

\newunicodechar{✓}{\ding{51}}
\newunicodechar{✗}{\ding{55}}

\title{
ViT-1.58b: Mobile Vision Transformers in the 1-bit Era}

\author{
 \textbf{Zhengqing Yuan\textsuperscript{1}\footnotemark[1]},
 \textbf{Rong Zhou\textsuperscript{2}\footnotemark[1]},
 \textbf{Hongyi Wang\textsuperscript{3,4}},
 \textbf{Lifang He\textsuperscript{2}},
 \textbf{Yanfang Ye\textsuperscript{1}},
 \textbf{Lichao Sun\textsuperscript{2}\footnotemark[2]}
\\
\\
 \textsuperscript{1}University of Notre Dame,
 \textsuperscript{2}Lehigh University,
 \textsuperscript{3}Carnegie Mellon University,
 \textsuperscript{4}Rutgers University
}

\begin{document}
\maketitle
\begin{abstract}

Vision Transformers (ViTs) have achieved remarkable performance in various image classification tasks by leveraging the attention mechanism to process image patches as tokens. However, the high computational and memory demands of ViTs pose significant challenges for deployment in resource-constrained environments. This paper introduces ViT-1.58b, a novel 1.58-bit quantized ViT model designed to drastically reduce memory and computational overhead while preserving competitive performance. ViT-1.58b employs ternary quantization, which refines the balance between efficiency and accuracy by constraining weights to \{-1, 0, 1\} and quantizing activations to 8-bit precision. Our approach ensures efficient scaling in terms of both memory and computation. Experiments on CIFAR-10 and ImageNet-1k demonstrate that ViT-1.58b maintains comparable accuracy to full-precision Vit, with significant reductions in memory usage and computational costs. This paper highlights the potential of extreme quantization techniques in developing sustainable AI solutions and contributes to the broader discourse on efficient model deployment in practical applications. Our code and weights are available at \url{https://github.com/DLYuanGod/ViT-1.58b}.

\end{abstract}

\renewcommand{\thefootnote}{\fnsymbol{footnote}} 
\footnotetext[1]{These authors contributed equally to this work.} 
\footnotetext[2]{Corresponding author.} 

\renewcommand{\thefootnote}{}

\footnotetext{Lichao Sun (lis221@lehigh.edu)}

\renewcommand{\thefootnote}{\arabic{footnote}}

\section{Introduction}

The rapid advancement of Transformer \cite{vaswani2017attention} architectures has significantly impacted the field of computer vision, particularly with the introduction of Vision Transformers (ViTs) \cite{dosovitskiy2020image}.
By treating image patches as tokens and leveraging the attention mechanism to process image patches, ViTs effectively capture complex dependencies across entire images, achieving remarkable performance in various image classification tasks \cite{graham2021levit, liu2021swin, yuan2021tokens}.
However, the computational and memory demands of ViTs are substantial, stemming primarily from their attention mechanisms, which scale quadratically with the number of tokens \cite{dosovitskiy2020image}. 
This inherent complexity poses significant challenges for deploying ViTs in resource-constrained environments such as mobile devices and embedded systems, where energy efficiency and low latency are critical

\begin{figure*}[!t]
\centering
\includegraphics[width=\textwidth]{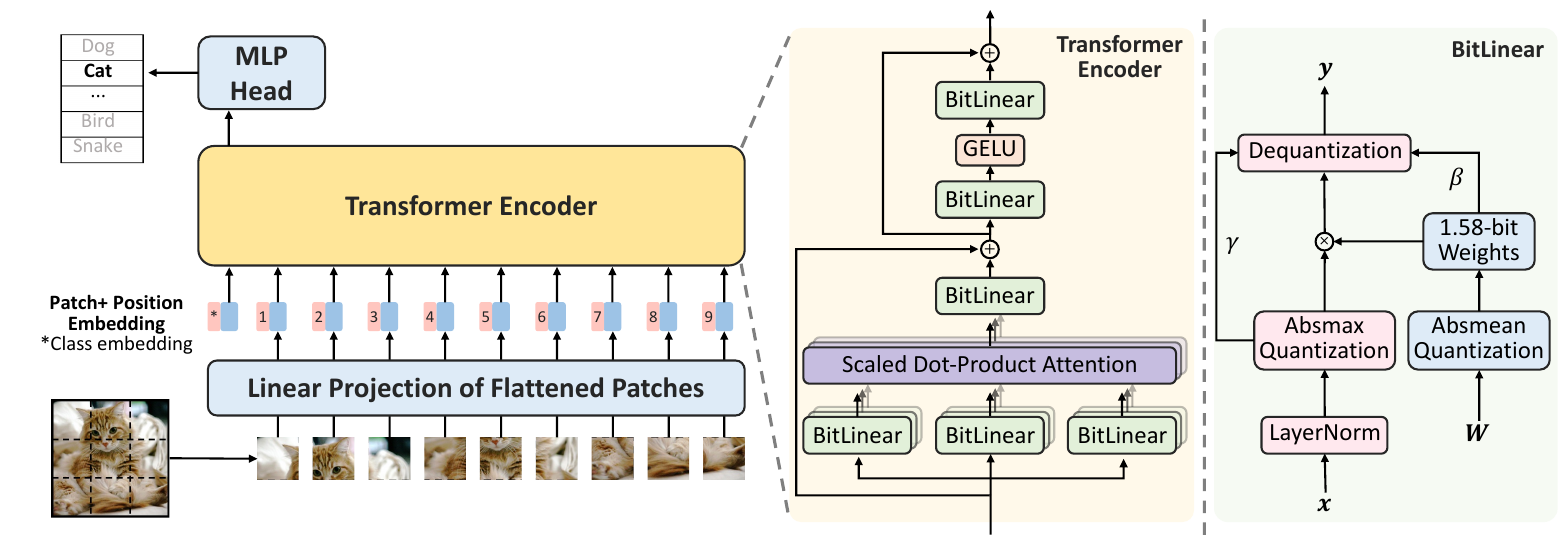} 
\caption{\textbf{An overview of the proposed ViT-1.58b:} ViT for image classification (left). Transformer encoder architecture (middle). BitLinear computation flow (right).}
\label{fig:frame}
\end{figure*}

Recently, research in neural network efficiency to mitigate these demands has explored various strategies including model pruning \cite{pan2023interpretability, yu2023unified, song2022cp}, knowledge distillation \cite{yang2022vitkd, touvron2021training, hao2022learning}, and quantization \cite{li2023vit, sun2022vaqf, lin2021fq}.
Among these, quantization techniques are particularly effective as they directly reduce the precision of weights and activations, thereby significantly lowering the memory and computational requirements of deep learning models.
Traditionally, post-training quantization has been favored for its simplicity as it does not necessitate changes to the training pipeline or retraining of the model. However, this method often results in significant accuracy loss at lower precision levels because the model isn’t optimized for the quantized representation during training \cite{frantar2022optq, chee2024quip}, limiting its utility for high-performance applications. 
In contrast, Quantization-Aware Training (QAT) \cite{park2018value} integrates the effects of quantization into the training process itself, simulating quantization effects to typically achieve better accuracy than post-training methods. For example, extreme quantization, like 1-bit models used in Binarized neural networks (BNNs) \cite{courbariaux2016binarized}, which utilize binary weights and activations, significantly decreasing computational and memory demands. Recent adaptations of 1-bit quantization techniques to Transformer models, such as BitNet \cite{wang2023bitnet}, and BiVit \cite{he2023bivit} have shown that even severe quantization can maintain performance while substantially cutting resource consumption.

However, these 1-bit models using binary quantization often face challenges in preserving model accuracy due to the extreme reduction in weight precision. To address this limitation, research has ventured into ternary quantization, which offers a more balanced approach. A notable example is "The Era of 1-bit LLMs: All Large Language Models are in 1.58 Bits" \cite{ma2024era}, which explores 1.58-bit quantization where weights can assume values of -1, 0, or 1. This approach refines the balance between performance and computational demands by introducing a zero value, which significantly reduces computational overhead. The inclusion of a non-polar weight (zero) not only allows for sparsity, thereby reducing the number of active computations but also maintains a richer representation than binary weights, potentially leading to less information loss. The success of this quantization strategy in large language models suggests its promising applicability to Vision Transformers, enabling more efficient processing of visual data while preserving model accuracy.

To address the unique demands of large-scale ViT models, we introduce ViT-1.58b, a 1.58-bit quantized ViT model. ViT-1.58b utilizes ternary quantization to optimize both memory and computational efficiency while maintaining competitive performance. This approach leverages the benefits of extreme quantization demonstrated in language models, adapting them to the context of computer vision. Our contributions are summarized as follows:
\begin{itemize} [nolistsep, leftmargin=*]
    \item We propose ViT-1.58b, the first 1.58-bit quantized ViT model, optimized for efficient scaling in terms of both memory and computation.
    \item We demonstrate the effectiveness of ViT-1.58b on the CIFAR-10 and ImageNet-1k datasets, showing competitive accuracy with significant reductions in memory and computational costs.
    \item We provide a comprehensive evaluation, comparing ViT-1.58b with state-of-the-art quantization methods and full-precision ViT models, highlighting its advantages in resource-constrained environments.
\end{itemize}


\section{Methods}~\label{method} 
As shown in Figure \ref{fig:frame}, our ViT-1.58b architecture is primarily based on ViT for image classification. 
The process begins by dividing the input image into patches, followed by applying a linear projection to the flattened patches. 
These patches then undergo patch and position embedding, as well as class embedding, before being fed into the transformer encoder. 
The output from the transformer encoder is processed through an MLP to produce the final classification prediction. 
We employ BitLinear layers from BitNet b1.58 \cite{ma2024era} to replace the conventional \textit{nn.Linear} layers in the transformer encoder. Different quantization functions are employed to quantize the weights to 1.58-bit precision and the activations to 8-bit precision, ensuring both efficiency and performance. Next, we will introduce how we implement BitLinear layers to achieve 1.58-bit weights and 8-bit activations.

\noindent\textbf{Weights Quantization}~~
In the ViT-1.58b, the 1.58-bit weights are achieved using the \textit{absmean} quantization function that constrains the weights to \{-1, 0, 1\}. 
Specifically, we first scale the weight matrix $W \in \mathbb{R}^{n \times m}$ by its average absolute value, thus obtaining $\frac{W}{\alpha+\epsilon}$, where \(\epsilon\) is a small floating-point number to avoid division by zero and ensure numerical stability, and \(\alpha\) is the average absolute value of the weight matrix, calculated as:
\begin{equation}
\alpha = \frac{1}{nm} \sum_{ij} |W_{ij}|.
\end{equation} 
Then, we round each value in the scaled matrix to the nearest integer among {-1, 0, +1}:
\begin{equation}
\tilde{W} = \text{RoundClip}\left(\frac{W}{\alpha+\epsilon}, -1, 1\right),
\end{equation}
\begin{equation}
\text{RoundClip}(x, a, b) = \max(a, \min(b, \text{round}(x))).
\end{equation}
This \textit{RoundClip} function rounds the input value \(x\) to the nearest integer and clips it within the specified range \([a, b]\).

Using this method, we can convert the weight matrix \(W\) into a ternary matrix $\tilde{W}$, where each weight value is constrained to \{-1, 0, 1\}.

\noindent\textbf{Activations Quantization.}~~
Following \cite{dettmers2022gpt3}, activations are quantized with b-bit precision by \textit{absmax} quantization function that scales the activations to the range \([-Q_b, Q_b]\), where \(Q_b = 2^{b-1}\). In the proposed ViT-1.58b, we set \(b = 8\), meaning that the activations are quantized to 8-bit precision, providing a balance between computational efficiency and maintaining sufficient precision for effective performance as described in \cite{ma2024era}. The activations quantization process is described as follows:
\begin{equation}
\tilde{x} = \text{Quant}(x) = \text{Clip}\left(x\times\frac{ Q_b}{\gamma}, -Q_b+\epsilon, Q_b-\epsilon\right),
\end{equation}
where \(\gamma = \max(|x|)\), representing the maximum absolute value of the activations, and the \textit{Clip} function ensures that the scaled values are confined within the specified range \([-Q_b, Q_b]\), defined as:

\begin{equation}
\text{Clip}(x, a, b) = \max(a, \min(b, x))
\end{equation}


The core component of ViT-1.58b is the BitLinear layer, which replaces the traditional \textit{nn.Linear} layer in the Transformer. Initially, the activations undergo a normalization layer \cite{ba2016layer} to ensure that the activation values have a variance of 1 and a mean of 0, mathematically represented as \(\text{LN}(x) = \frac{x - \mathbb{E}(x)}{\sqrt{\text{Var}(x) + \epsilon}}\). Following this, normalized activations are quantized using the \textit{absmax} quantization function, resulting in \(\text{Quant}(\text{LN}(x))\). With weights quantized to 1.58-bit precision, matrix multiplication results in the output \(y = \tilde{W} \cdot \text{Quant}(\text{LN}(x))\), and the output is subsequently dequantized to rescaled with \(\beta\) and \(\gamma\) to the original precision, expressed as \(y_{dequant} = y \times \frac{\beta \gamma}{Q_b}\), where \(\beta = \frac{1}{nm} \|W\|_1\)

These steps collectively enable the \textit{BitLinear} layer to maintain the performance of the Transformer model while significantly reducing computational cost and storage requirements.

\noindent\textbf{Training Strategy}~~
During training, we employ the Straight-Through Estimator (STE) \cite{bengio2013estimating} to handle non-differentiable functions such as the sign and clip functions in the backward pass. The STE allows gradients to flow through these non-differentiable functions, enabling effective training of the quantized model.

Additionally, we use mixed precision training, where weights and activations are quantized to low precision, but gradients and optimizer states are kept in high precision to ensure training stability and accuracy.

\section{Experiments and Results}~\label{experiment}
\noindent\textbf{Experimental Setting.}~~
We evaluate our ViT-1.58b model on two datasets, CIFAR-10 \cite{krizhevsky2009learning} and ImageNet-1k \cite{deng2009imagenet}, comparing it with several versions of the Vision Transformer Large (ViT-L): the full-precision ViT-L (i.e. 32-bit precision), and the 16-bit, 8-bit, and 4-bit inference versions in terms of memory cost, training loss, test accuracy for CIFAR-10, and Top-1 and Top-3 accuracy for ImageNet-1k.

The computational framework for this study comprised four NVIDIA TESLA A100 GPUs, each with 80 GB of VRAM. The system's processing core utilized an AMD EPYC 7552 48-core Processor with 80 GB of system RAM to manage extensive datasets efficiently. We employed PyTorch version 2.0.0 integrated with CUDA 11.8, optimizing tensor operations across GPUs.

\begin{table}[ht]
\centering
\resizebox{\columnwidth}{!}
{
\begin{tabular}{@{}l|c|cc|cc@{}}
\toprule
\multirow{2}{*}{Method} & \multirow{2}{*}{Memory Cost} & \multicolumn{2}{c|}{CIFAR-10} & \multicolumn{2}{c}{ImageNet-1k}  \\ \cmidrule(lr){3-6}
                       & & train loss   & test acc. & Top-1 & Top-3   \\ \midrule
ViT-L  &1.14G  & 0.024  & 76.28 & 76.54 & 90.23 \\ 
16bit Inference ViT-L & 585M   & -  & 74.61 & 75.29 & 87.44 \\
8bit Inference ViT-L   &293M & -  & 72.20 & 74.11 & 85.26 \\
4bit Inference ViT-L   &146M & -  & 70.69 & 70.88 & 82.50 \\ \midrule
ViT-1.58b-L (ours)          &57M & 0.026  & 72.27 & 74.25 & 85.78 \\
\bottomrule
\end{tabular}
}
\caption{This table presents a comparative performance of various versions of the ViT-L model on the CIFAR-10 and ImageNet-1k datasets. 
}
\label{tab:rec-ablation}
\end{table}

\noindent\textbf{Results.}~~
Table~\ref{tab:rec-ablation} shows the performance of the ViT-L model across different bit precisions evaluated on two widely recognized datasets: CIFAR-10 and ImageNet-1k. The results highlight how the reduction in bit precision from full precision (ViT-L) to lower bit configurations (16-bit, 8-bit, 4-bit, and 1.58-bit) affects the model's effectiveness in terms of training loss, test accuracy, and top-k accuracy metrics. 

\begin{figure}
\centering
\includegraphics[width=0.5\textwidth]{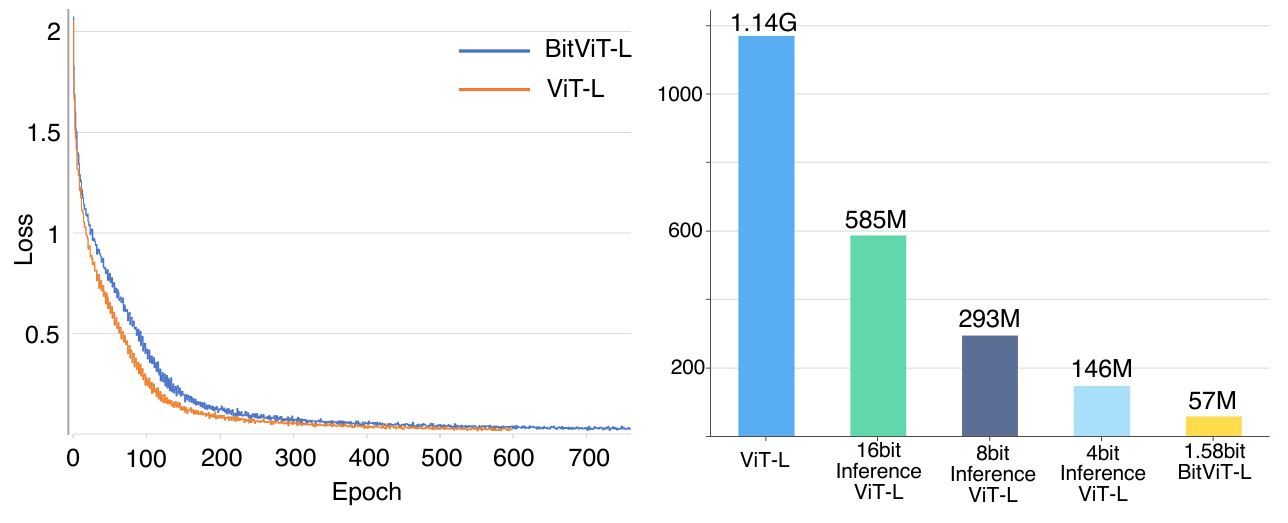} 
\caption{Training loss curve (left). memory consumption of each model (right).}
\label{fig:f2}
\end{figure}

For CIFAR-10, the full-precision ViT-L model achieved a test accuracy of 76.28\%, which serves as a baseline for comparison. When the bit precision was reduced to 16-bit, there was a moderate decline in accuracy to 74.61\%. Further reduction to 8-bit and 4-bit resulted in more pronounced declines to 72.20\% and 70.69\%, respectively. This trend suggests that lower bit precision generally degrades the model's performance, likely due to the increased quantization error and reduced capacity to capture the variability in the data. The proposed 1.58-ViT-1.58b-L model, which operates at an even lower bit precision than the other quantized variants, recorded a test accuracy of 72.27\%. Interestingly, the performance of this model is closer to the 8-bit version than to the 4-bit.

On the more complex and diverse ImageNet-1k dataset, a similar pattern is observed. The full-precision ViT-L model achieved a Top-1 accuracy of 76.54\% and a Top-3 accuracy of 90.23\%. Reduction to 16-bit precision caused declines to 75.29\% and 87.44\% in Top-1 and Top-3 accuracies, respectively. These declines became more substantial at 8-bit (Top-1: 74.11\%, Top-3: 85.26\%) and 4-bit (Top-1: 70.88\%, Top-3: 82.50\%). The 1.58-bit model managed a Top-1 accuracy of 74.25\% and a Top-3 accuracy of 85.78\%, showcasing a performance that surpasses the 8-bit version in Top-1 accuracy and nearly matches it in Top-3 accuracy. 


As shown in Figure~\ref{fig:f2}, ViT-1.58b-L shows promising results in both training loss and memory consumption compared to the full-precision ViT-L and its quantized versions. The left panel depicts training loss curves, where ViT-1.58b-L's loss closely follows that of the full-precision ViT-L. This suggests that the 1.58-bit quantization retains the model's learning capacity effectively. The right panel highlights the memory consumption. The full-precision ViT-L requires 1.14 GB of memory, while ViT-1.58b-L drastically reduces this to just 57 MB. This significant reduction in memory usage makes our model ideal for deployment in resource-constrained environments. Overall, our ViT-1.58b-L model balances competitive performance with substantial memory savings, demonstrating its efficiency and practicality for real-world applications.

\begin{figure}
\centering
\includegraphics[width=0.4\textwidth]{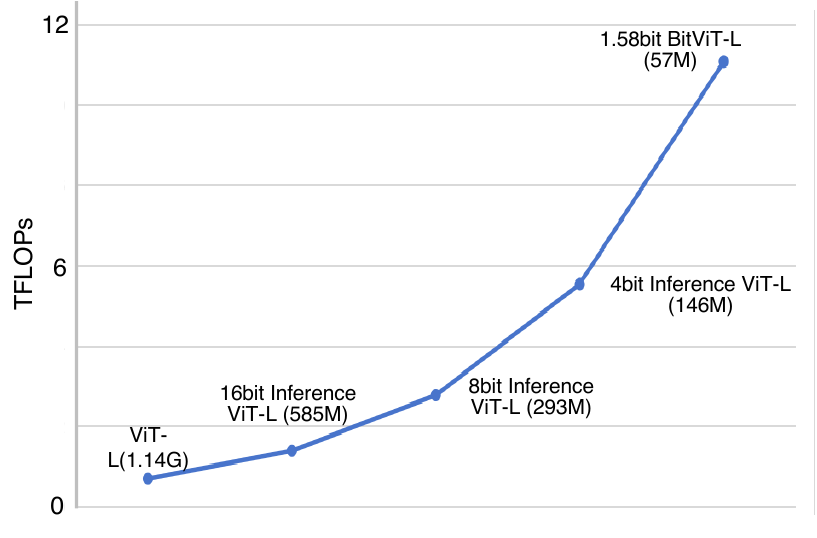} 
\caption{Computational performance in TFLOPs for different bit precisions.}
\label{fig:f3}
\end{figure}

In Figure~\ref{fig:f3}, as bit precision decreases, computational performance (measured in TFLOPs) significantly increases. Full-precision achieves 0.692 TFLOPs, while ours reaches 11.069 TFLOPs. This substantial increase in performance with lower precision highlights the efficiency gains achievable through extreme quantization. Our model offers a dramatic boost in computational throughput, making it highly suitable for high-performance computing environments where both speed and resource efficiency are critical.


\section{Conclusion}~\label{conclu}
In this paper, we introduced ViT-1.58b, a 1.58-bit quantized Vision Transformer that efficiently addresses computational and memory challenges in vision model deployment through ternary quantization. Our results show that ViT-1.58b achieves competitive accuracy on benchmarks like CIFAR-10 and ImageNet-1k with significantly lower resource requirements. This model demonstrates the potential of advanced quantization techniques in complex Transformer architectures, highlighting their role in developing sustainable AI solutions. Future work will explore ViT-1.58b's scalability and integration into larger systems, enhancing its practical utility and environmental benefits.

\section{Limitations}
While the ViT-1.58b model exhibits promising performance and demonstrates efficient computational usage, it also presents certain limitations that need to be addressed. This section outlines the primary challenges associated with our approach.

\noindent\textbf{Requirement for Pre-Training.}~~
One significant limitation of the ViT-1.58b architecture is the necessity for pre-training the ViT model from scratch when applying our ternary quantization techniques. This requirement can significantly raise the barrier to entry for utilizing our proposed model, as pre-training demands extensive computational resources and time. Organizations or individuals with limited access to such resources might find it challenging to adopt this technology without pre-trained models readily available. Moreover, pre-training introduces additional complexity in ensuring the robustness and generalizability of the model before it can be deployed effectively.

\noindent\textbf{Training Difficulty.}~~
Another critical challenge is the increased difficulty in training the 1.58-bit ViT model compared to its full-precision counterpart. During our experiments, as shown in Figure~\ref{fig:f2}, we observed that training the 1.58-bit ViT on CIFAR-10 required approximately 250 epochs to achieve a training loss around 0.026, whereas the standard ViT could reach a comparable loss of 0.024 in just 200 epochs. This increased training duration for the 1.58-bit model suggests a lower learning efficiency, likely due to the reduced precision and the consequent limitations in the model's capacity to capture detailed feature representations.


\bibliography{references}

\end{document}